\documentclass{article}

\usepackage{amsmath, tikz}
\usepackage{framed}
\usetikzlibrary{bayesnet}

\usepackage{times}
\usepackage{graphicx} 
\usepackage{subfigure} 

\usepackage{natbib}

\usepackage{algorithm}
\usepackage{algorithmic}

\usepackage{hyperref}

\usepackage[accepted]{icml2015}

\icmltitlerunning{Fast Latent Variable Models for Visualization on Mobile Devices}

\begin{document} 

\twocolumn[
\icmltitle{Fast Latent Variable Models for Inference and Visualization on Mobile Devices}

\icmlauthor{Joseph W Robinson}{jwrobins@andrew.cmu.edu}
\icmladdress{School of Computer Science,
            Carnegie Mellon University. Pittsburgh, PA, 15213}
\icmlauthor{Aaron Q Li}{aaron@potatos.io}
\icmladdress{School of Computer Science,
            Carnegie Mellon University. Pittsburgh, PA, 15213}

\icmlkeywords{boring formatting information, machine learning, ICML}

\vskip 0.3in
]

\begin{abstract} 
In this project we outline Vedalia, a high performance distributed network for performing inference on latent variable models in the context of Amazon review visualization. We introduce a new model, RLDA, which extends Latent Dirichlet Allocation (LDA) \cite{OriginalLDA2003} for the review space by incorporating auxiliary data available in online reviews to improve modeling while simultaneously remaining compatible with pre-existing fast sampling techniques such as \cite{SparseLDA2009, LietalKDD2014} to achieve high performance. The network is designed such that computation is efficiently offloaded to the client devices using the Chital system \cite{896Project}, improving response times and reducing server costs. The resulting system is able to rapidly compute a large number of specialized latent variable models while requiring minimal server resources.
\end{abstract}

\section{Introduction}
Recent advances in machine learning have given rise to several classes of latent variable models, allowing one to embed additional unobserved structure into a problem in order to improve results. One such method is Latent Dirichlet Allocation (LDA) \cite{OriginalLDA2003}, a generative model in which documents are assumed to contain a mixture of topics; each topic is then represented as a probability distribution over words in the corpus. A recent application of this is in Amazon review modeling \cite{715Project}. In this approach, each review text is treated as a document, and products are displayed via a word cloud containing the top $n$ words per topic. However, while work such as the Alias method \cite{LietalKDD2014}, GPU-sampling \cite{LietalANU2012}, and Parameter Server \cite{LietalNIPS2013, LietalWSDM2015} has resulted in substantial speed improvements for large scale systems, much less has been done for a more hardware-constrained setting such as smartphones running Android or iOS. Performance aside, even less has been explored with respect to obtaining interpretable and visualizable results from these models.

\section{Background}
\subsection{Current Amazon System}
The current Amazon system for displaying reviews is divided into three sections. Quotes give a user a set of three one-line excerpts from reviews as a high level overview of the product. A list of ``most helpful reviews'' give the user more detailed information via the eyes of a select few reviewers. The full review list is then available as a backup for the ambitious buyer who wishes to sift through thousands of reviews to ensure their purchase will be worthwhile. It is easy to spot a couple potential flaws in this system -- a user is either able to view the experience of a small choice set of individuals, or they must look through the full uncurated review set. Thus, their is no aggregation of knowledge despite the availability of millions of reviews throughout the site. In addition, the system does not quickly convey a multi-faceted view of the product. For example, a user buying a smartphone might be interested in knowing the general sentiment about the phone's camera, battery life, performance (does it lag?), reliability of connection, etc.

\subsection{Previous Topic Modeling Approach}
The previous approach uses LDA to overcome this issue by representing each product by its topic distribution. Each topic distribution is then displayed to the user via a word cloud in order to provide multifaceted information about a product in a way that is intuitive, compact, and aggregate. The system retains completeness in the sense that the full review set is still available to the user via an interactive topic-based search feature. However, this system still falls short in several areas. Namely, the system:
\begin{itemize}
\item Requires that models are built on the cloud and returned for the user for display
\item Is far too slow to be run on mobile operating systems
\item Does not take advantage of auxiliary data such as ratings, helpfulness votes, and users when estimating topics
\item Utilizes a fixed number of topics (16) regardless of the product
\item Does not give rise to results that are easily displayable on a small-screen mobile device
\item Uses a fixed dataset \cite{amazondata} rather than an adaptive one
\end{itemize}

The first point gives rise to several issues when scaling the service. Previously, computing topics was done with MALLET \cite{MALLET}, a machine learning toolkit written in Java. Aside from the toolkit's inability to efficient parallelize LDA training, the use of MALLET server-side results in large computation time and an unscalable system in terms of cost. We plan to address this issue by building a system that can run efficiently in a mobile setting, thus offloading the vast majority of computation to a distributed system of clients connected by a central model cache and updating server. To do this, the proposed learning algorithm must efficiently run in a locally parallel system (2-4 cores) in an amount of time that is acceptable for a typical user. Sampling for the algorithm must be done from scratch; as an example of MALLET's poor performance on Android, it was found that a small topic modeling job (~350 reviews) caused the app to crash after several minutes largely spent in garbage collection when using MALLET due to high memory consumption.

Another weakness of the previous system is omission of auxiliary data from the underlying probability model. By using standard LDA, ratings, helpfulness votes, and the user network cannot be used for directly improving results. While post-processing can help with some of these issues, it cannot effectively improve the quality of topics themselves. Thus, we present a new latent variable model, RLDA, that incorporates this information in order to improve topic quality and reduce noise. In addition, RLDA allows for a variable number of topics in order to help avoid the display of information-void topics and improve user experience.

Along these same lines, the final result must be easily visualizable on a mobile device. While the previous system was easily viewable on a desktop web browser, the size constraints of a mobile device coupled with the lack of mouse prohibit the use of many previous features. Namely, the number of topics displayed and the size of the visualization circle are too large to display on a mobile device while maintaining visible text, and the lack of a mouse invalidates the hover-based review selection system. In later sections, we demonstrate how our revised approach overcomes these issues to allow RLDA results to effectively be visualized on small-screen devices.

As a final note, the previous system utilized a fixed dataset \cite{amazondata} which did not include recent products or allow for model updating as new reviews appear. For a potential buyer, it is often useful to know if a product has a tendency to fail a few months after purchasing it. We address this issue by creating a system that dynamically updates models as new data becomes available using efficient sampling techniques.

\subsection{Pre-existing work on modeling reviews}
Pre-existing latent variable models designed for analyzing reviews, such as \cite{BrodyE10, JoO11, TitovM08}, generally fall short in scalability and generality. They generally make improvement over LDA by using word associations and sentence context to form more representative words. In addition to that, they also focus specifically on a fixed number of known aspects. This severely limits the potential application of these models, and as a result they cannot capture the unknown topics/aspects at fine-grained level (e.g a third-party charging adapter does not work with some Apple computers). These models also ignore a large quantity of auxiliary data such as user ratings, helpfulness, and unhelpfulness. 

\subsection{Latent Dirichlet Allocation}
Latent Dirichlet Allocation (LDA) \cite{OriginalLDA2003} is a widely used topic model in which documents are assumed to be generated from mixture distributions of language models associated with individual topics. That is,  the documents are generated by the latent variable model below:
\begin{figure}[ht!]
\centering
\begin{tikzpicture}

    \node[obs]                              (alpha) {$\alpha$};
    \node[latent, right=of alpha]                               (theta) {$\theta_d$};
    \node[latent, right=of theta]                               (z) {$z_{di}$};
    \node[obs, right=of z]                               (w) {$w_{di}$};
    \node[latent, right=of w]                               (phi) {$\phi_k$};
    \node[obs, right=of phi]                               (beta) {$\beta$};

    \edge{alpha}{theta} ;
    \edge{theta}{z} ;
    \edge{z, phi}{w} ;
    \edge{beta}{phi} ;
    
    \plate {plate1} {(z)(w)} {for all $i$} ;
    \plate {plate2} {(theta)(plate1)} {for all $d$} ;
    \plate {plate3} {(phi)} {for all $k$} ;

\end{tikzpicture}
\end{figure}
\\ The generative process is as follows: \\
\\
For each document $d$ draw a topic distribution $\theta_d$ from a Dirichlet distribution with parameter $\alpha$
\begin{align}
\theta_d \sim Dir(\alpha)
\end{align}
For each topic $t$ draw a word distribution from a Dirichlet distribution with parameter $\beta$
\begin{align}
\psi_t \sim Dir(\beta)
\end{align}
For  each word $i \in \{1...n_d\}$ in document $d$ draw a topic from the multinomial $\theta_d$ via
\begin{align}
z_{di} \sim Mult(\theta_d)
\end{align}
Draw a word from the multinomial $\psi_{z_{di}}$ via
\begin{align}
w_{di} \sim Mult(\psi_{z_{di}})
\end{align}
The Dirichlet-multinomial design in this model makes it simple to do inference due to distribution conjugacy -- we can integrate out the multinomial parameters $\theta_d$ and $\psi_k$, thus allowing one to express $p(w,z|\alpha,\beta,n_d)$ in a closed-form \cite{SparseLDA2009}. This yields a Gibbs sampler for drawing $p(z_{di}|rest)$ efficiently. The conditional probability is given by
\begin{align}
p(z_{di}|rest) \propto \frac{(n_{td}^{-di} + \alpha_t)(n_{tw}^{-di} + \beta_w)}{n_t^{-di} + \bar{\beta}}
\end{align}
Here the count variables $n_{td},n_{tw}$ and $n_t$ denote the number of occurrences of a particular (topic,document) and (topic,word) pair, or of a particular topic, respectively. Moreover, the superscript $.^{-di}$ denotes count when ignoring the pair $(z_{di},w_{di})$. For instance, $n_{tw}^{-di}$ is obtained when ignoring the (topic,word) combination at position $(d,i)$. Finally, $\bar{\beta}:=\sum_{w}\beta_w$ denotes the joint normalization.

Sampling from (5) requires $O(k)$ time since we have $k$ nonzero terms in a sum that need to be normalized. In large datasets where the number of topics may be large, this is computationally costly. However, there are many approaches for substantially accelerating sampling speed by exploiting the topic sparsity to reduce time complexity to $O(k_d + k_w)$ \cite{SparseLDA2009} and further to $O(k_d)$ \cite{LietalKDD2014}, where $O(k_d)$ denotes the number of topics instantiated in a document and $O(k_w)$ denotes the number of topics instantiated for a word across all documents.
\subsection{Chital Computation Marketplace}
Chital is a scalable, distributed computation marketplace designed for the efficient allocation of high CPU, low network bandwidth tasks among a network of mobile devices. The five key aspects of Chital are 1) task distribution via the marketplace, 2) a credit score system for monitoring user behavior 3) real-time matching mechanisms for maximizing user gain 4) an optional lottery system for further incentivizing participation 5) an evaluation system for verifying the submitted models. Each of these is discussed in detail below.
\subsubsection{Marketplace}
\label{Marketplace}
The marketplace is the major underlying component in Chital for task allocation. In the marketplace, each user has the option of opting-in to background computation; once opted in, this user is then listed as a computational seller and can be assigned modeling tasks to be run in the background. When a Vedalia user enters a query, a matching request is sent to a centralized server system -- this user is now a buyer. Assuming the buyer has sufficient computational power on his phone, the buyer is also automatically listed as a seller for the duration of his model computation. The marketplace then matches the buyer with a pair of sellers and requests that both sellers generate a model from the supplied data. This data is then returned to the central servers, where the system determines whether model verification is necessary. Let $c_1$ and $c_2$ 
denote the credit of the two sellers, and $p_1$ and $p_2$ denote the 
perplexity of the sellers' results. Then the probability of secondary verification 
is defined as:
\begin{eqnarray}
1 - \frac{1}{3} \Bigg[ \frac{1}{1 + e^{-(c_1 + c_2)}} + 2\frac{\min(p_1, p_2)}{\max(p_1, p_2)} \Bigg]
\end{eqnarray}
Thus, high seller credit scores and high perplexity match reduce the probability of verification, and vice versa. The best model  (measured by perplexity) that passes verification is then returned to the original buyer.
\subsubsection{Credit System}
A 0-sum credit system is established that begins with two 0-credit sellers for computation. Then, each user that joins the system as a seller begins with 0 credit. When building a model, the perplexities of each of the two models returned by the sellers are compared; a credit from the worst model's seller is then transferred to the best model's seller. The best model's seller is additionally awarded a $t \cdot i^*$ lottery tickets, where $t$ is the number of tokens processed and $i^*$ is the number of sampling iterations performed by the best model. Assuming every seller is honest, each seller has expectation 0 credit over time. However, in the event that a malicious seller attempts to provide phony results in order to acquire lottery tickets, the credit distribution shifts from the bad to good users. As a result, the system becomes less likely to need to verify results of good users, and becomes increasingly likely to perform verification on bad users.

\subsubsection{Matching Mechanisms}
The core of Chital is a real-time matching system that pairs each query with two sellers. The matching problem can be formulated as a bipartite matching problem with both sides of the vertices arriving online. Each buyer vertex is required to match with two seller vertices. In addition to that, after a matching is established, the matched vertices only become temporarily unavailable for a period of time based on the performance of seller nodes and the task size of buyer node, before the matching is removed and the vertices become available again.

Although online graph matching especially online bipartite matching is a well-studied major research area \cite{KarVazVaz90, Mehta13}, our problem setup makes it difficult to apply any existing algorithm for two reasons: our problem introduces an extra "time dimension", and our objective is to maximize overall user gain thereby to convince them joining the system voluntarily. Based on these, we developed the concept of strategyproofness and Nash equilibrium in another recent work of ours \cite{896Project} and, studied and created a suite of new real-time matching algorithms to achieve our goal.

\subsubsection{Lottery System}
To further incentivize seller participation, a lottery system can be constructed in which a portion of app advertising revenue is allocated for a lottery system. At the end of each lottery period, a user a sampled at random with probability of winning proportionate to the user's number of lottery tickets. The full lottery amount is then awarded to this winning user.

Note, however, the existence of the lottery system is entirely optional, since a rational user would voluntarily participate the system if a good matching mechanism with strategyproofness and empirical Nash equilibrium were used. In our empirical studies \cite{896Project}, we found under appropriate parameters, users always save overall computation time by a large margin within our simulation.

\subsubsection{Evaluation System}
Evaluation is a multi-stage system consisting of model validation, selection, and verification.

In validation, basic properties of the submitted distributions are verified (e.g. sum to 1). Any model that fails validation is immediately rejected.

In selection, the perplexity of each submitted model is computed. The model with the lower perplexity is the selected to be returned to the end user, pending verification.

In verification, the marketplace first computes the probability $p_v$ of secondary verification as described in \ref{Marketplace}. A value $s$ is then sampled uniformly from [0,1]; if $s > p_v$, verification occurs. In verification, a few additional iterations of Gibbs sampling are run on the selected model on Chital servers, and final model perplexity is computed. If the final perplexity deviates substantially from that of the submitted model, the submitted model has not converged and is thus rejected by the system.
\section{Proposed System}
We propose a system that incorporates a new latent variable model, RLDA, that is well-suited for mobile devices and review analysis. It naturally extends LDA while simultaneously maintains a structure that allows the use techniques introduced in SparseLDA and AliasLDA to achieve high-performance at any scale. Furthermore, the system is integrated with Chital for scalability, and is accessible to the end user in the form of a mobile application.

\subsection{RLDA}
Review-augmented Latent Dirichlet Allocation (RLDA) is an adaptation of LDA that is well-suited for modeling reviews in a mobile setting due to its high sampling performance and increased structure with respect to standard LDA. In Figure \ref{RLDA-model} we present the RLDA model in plate notation. The notations are described in Figure \ref{RLDA-vars}
\begin{figure}[ht!]
\label{RLDA-model}
\centering
\begin{tikzpicture}

    \node[obs]                              (alpha) {$\alpha$};
    \node[latent, right=of alpha, xshift=6mm]                               (theta) {$\theta_d$};
        \node[latent, above=of theta]                               (rcat) {$c_d$};
    \node[latent, above=of rcat]                               (radj) {$\widetilde{r}_d$};
    \node[obs, above=of radj, xshift=-12mm]                               (r) {$r_d$};
    \node[obs, above=of radj]                               (b) {$b_d$};
    \node[obs, above=of radj, xshift=12mm]                               (sigma) {$\sigma^2_d$};
    \node[latent, below=of theta]                               (z) {$z_{di}$};
    \node[obs, below=of z]                               (w) {$w_{di}$};
    \node[latent, right=of w, xshift=6mm]                               (phi) {$\phi_{dk}$};
    \node[latent, above=of phi]                               (psi) {$\psi_d$};
    \node[obs, above=of psi, xshift=12mm]                               (h) {$h_d$};
    \node[obs, above=of psi]                               (u) {$u_d$};
    \node[obs, above=of psi, xshift=-12mm]                               (nu) {$\nu_d$};
    \node[obs, right=of phi, xshift=6mm]                               (beta) {$\beta$};

    \edge{alpha, rcat}{theta} ;
    \edge{radj}{rcat} ;
    \edge{r, sigma, b}{radj} ;
    \edge{theta}{z} ;
    \edge{z, phi}{w} ;
    \edge{beta, psi}{phi} ;
    \edge{u, h, nu}{psi} ;
    
    \plate {plate1} {(z)(w)} {for all $i$} ;
    \plate {plate2} {(phi)} {for all $k$} ;
    \plate {plate3} {(theta)(radj)(r)(sigma)(h)(u)(nu)(b)(plate1)(plate2)} {for all $d$} ;

\end{tikzpicture}
\caption{RLDA}
\end{figure}\\
\begin{figure}[h!]
\label{RLDA-vars}
\begin{framed}

  $r_d$ : rating associated with review $d$ \\
  $b_d$ : mean of user $d$'s rating biases, excl. review $d$\\
  $\sigma^2_d$ : variance of user $d$'s rating biases, excl. review $d$\\
  $\widetilde{r}_d \sim N(r_d + b_d, \sigma^2_d + 1)$ : bias-corrected review rating \\
  $c_d$ : categorical distribution over ratings ${1, 2, 3, 4, 5}$ \\
  $\nu_d$ : writing quality score for review $d$ \\
  $u_d$ : unhelpfulness votes for review $d$ \\
  $h_d$ : helpfulness votes for review $d$ \\
  $\psi_d \sim Bernoulli(Logistic(\nu_d, u_d, h_d))$ : the review quality rating 

\end{framed}
\caption{Variables in RLDA}
\end{figure}

Note that despite the introduction of latent variables $\widetilde{r}_d$, $c_d$, $\psi_d$, and our per-document observed variables $r_d$, $\sigma_d$, $h_d$, $u_d$, $\nu_d$, the basic structure of LDA is maintained. However, we can see that the topic distribution of each review is now dependent on the review's bias-corrected rating. This makes sense intuitively in that we expect more negative reviews to talk about different topics that wholly positive ones; as an example, negative reviews might tend to focus on poor product quality and customer service, with positive reviews focusing on product satisfaction and example use cases. The model also incorporates a Bernoulli review quality rating $\psi_d$, which takes into account review helpfulness votes, unhelpfulness votes, and writing quality (out-of-vocabulary rate, punctuational correctness, average word length, etc.). 

Creating a model while simultaneously maintaining high sampling performance can be very challenging. To our knowledge, many pre-existing latent variable models overlook this issue albeit providing interesting results in accuracy in a laboratory setting for certain categories of reviews. In Section \ref{SamplingSection}, we describe efficient sampling techniques for the RLDA model that build on existing LDA sampling methods.

\subsection{Model Updating}
Using the Chital system, model updating follows naturally by performing sampling using the existing model with the new reviews added to the review set. In this way, if a lottery system is used, the number of lottery tickets awarded to the seller is fairly determined by the amount of computation required to update the model. To avoid convergence to poor optima, we recompute a product model after every few updates. This methodology allows for products to be quickly updated when new reviews become available while maintaining model quality via occasional full recomputes.

\subsection{Core Set Selection}
To accommodate a variable number of topics, we first perform RLDA sampling with a fixed number of topics $k$. The number of topics can then be reduced to a smaller core set post-sampling by using techniques in \cite{FeldmanFK11} combined with estimating the informativeness of the top words in each topic.
\subsection{Visualization}
Given the limited screen space available on mobile devices, the interface is designed with simplicity in mind from the ground up. The initial screen is a simple search, with a singular entry box in which the user can query products. After submitting a product query, the user is provided with a list of products to select from in order to build a topic model.

In contrast to \cite{715Project}, the display of topics is returned to its core. We display each topic using its review in topic-document sorted order, however in contrast to Amazon's system we partition the visualization in a set of tabs, one for each topic. The user can the use an intuitive SeekBar to select topics. Once selected, a topic summary is displayed including topic weighted rating, topic weight, and the top $k$ tokens of the topic listed as keywords. Above this topic summary is a a review ViewPager, which can be using to quickly pan through reviews in sorted order according to topic probability of the selected topic. Within the review text, each region of text corresponding to a keyword lemma is bolded in order to bring attention to regions of the review that are pertinent to the selected topic. In this way, I used can quickly glance at select regions of a review when learning about features of a product that they deem important. We defer more discussion of visualization to the case studies in which we provide examples of Quokka visualizations.
\section{Implementation Details}
Below we briefly describe the implementation of Vedalia, with emphasis on performance-critical details and modeling.

\subsection{Architecture, Preprocessing, and Database}
To accommodate the new system design we made substantial architecture change compared to the previous system \cite{LiRobinsonDengJing14}. We dropped the integration with parameter-server-like computing clusters, but substantially increased the power of pre-processing clusters and databases. Furthermore, the model result selection system and intermediate-result push-update mechanism are no longer required, hence entirely replaced by Chital system.

We schedule large-scale batch review preprocessing task using Apache Spark \cite{spark10} combined with Stanford CoreNLP \cite{stanfordnlp14} as soon as enough reviews are committed into the database, so to reduce waiting time and eliminate overhead during model computation. We deployed a 1-rack, multi-node Cassandra \cite{cassandra10} database for storing and streaming product information, raw reviews, and analyzed reviews, so to achieve best performance in fault tolerance, consistency, and read-and-write at scale. An analyzed review is a review attached with pre-processing result in compressed binary format. 

To facilitate searching we deployed a multi-node scalable search engine, ElasticSearch \cite{elasticsearch15}, that operates alongside Cassandra, simultaneously indexes every product and every review inserted to the database. ElasticSearch complements Cassandra with its higher insertion performance and much more flexible query structure, while simultaneously taking benefit from Cassandra's high consistency and fault tolerance. Web servers are deployed in the front end which exposes APIs that process product and review queries and stream the result back to query initiators, while simultaneously provides network-level isolation and security guarantee.

This architecture effectively establishes a scalable system which is not sensitive to the number of users making queries, due to Chital's ability to offload computation to users themselves. Since reviews are preprocessed in batch at insertion as Spark tasks, large amount of workers are only allocated temporarily for a short period of time. Modern cloud computing platforms such as Google Compute Platform allow this to be done in an automated fashion, and charge for only the amount of time allocated on a per-minute basis. To process the entire collection of 23 million reviews in SNAP dataset \cite{amazondata}, we used only up to 100 cores with 500GB of memory resources for 2 days.
\subsection{Model Views}
To reduce bandwidth and protect models from outside use, we avoid sending the entire model to the end user. The initial model view is streamed to the user as a list of topic descriptions (id, probability, expected rating, expected helpfulness, expected unhelpfulness) and their associated top $n$ words. As with the previous system, we defer sending review text to the end user until it is requested. This is of particular importance in a mobile setting, where many users will be using the app on a bandwidth-limited data plan. To improve user experience, reviews can be cached for offline viewing.
\subsection{Sampling}
\label{SamplingSection}
Sampling can be performed by following a procedure which transforms the auxiliary information along with other latent variables into word observation, then sample the transformed data in an LDA-like fashion, where an adaptation of SparseLDA \cite{SparseLDA2009} sampling is performed in order to estimate model parameters. We define review score tier $c_{d,t}$ as:
\begin{align*}
  c_{d,1} &:= p(\widetilde{r}_d \leq 1.5), \\
  c_{d,2} &:= p(\widetilde{r}_d \in (1.5, 2.5]), \\
  c_{d,3} &:= p(\widetilde{r}_d \in (2.5, 3.5]), \\
  c_{d,4} &:= p(\widetilde{r}_d \in (3.5, 4.5]), \\
  c_{d,5} &:= p(\widetilde{r}_d > 4.5), \\
\end{align*}

Additionally, we need to characterize the distribution of $\psi_d$. We train a logistic regression model mapping $\{ \nu_d, u_d, h_d \} \rightarrow \text{is\_relevant}$, where $\text{is\_relevant} = 1$ if the review is relevant to the product being reviewed, and 0 otherwise. As an example, one Amazon review for a Macbook Pro says "The product is good but I find that my neck is getting sore from using it."; the goal of the logistic model, then, is to label with review as not relevant. While the original intent was to train a model using data collected from Amazon Mechanical Turk, we later chose to hand-label a set of reviews in order to train our classifier as a means of cutting our implementation costs.

Since the vast majority of Amazon reviews come from users who have only reviewed a single product, estimating the distribution of a general user's bias-corrected rating is often impossible. In order to reduce within-topic rating variability, for a general user we assume low rating variance and approximate the rating distribution by adding the review only for the given rating. This is achieved by appending ``\_rating'' to each token within a review, then stripping out the rating suffix when displaying keywords to the user. Notice that in making this approximation we simultaneously utilize the imposed independence assumption that $\psi_d \perp c_d | w_{d*}$, as shown in the RLDA graphical model.

Approximate weighting is performed by allocating the bottom $w_{bits}$ bits of review-topic and word-topic counts for fractional counts. What previously would correspond to a count increment of 1 is mapped to an increment of $2^{w_{bits} + 1}$. Fractional counts can then be approximated as an integer-rounded fraction of $2^{w_{bits} + 1}$, providing us with $\frac{1}{2^{w_{bits} + 1}}$ precision. Count sparsity can be imposed by reducing the value of $w_{bits}$ -- all fractional counts below $\frac{1}{2^{w_{bits} + 2}}$ will be treated as a 0-count.
\section{Case Study}
In order to evaluate the effectiveness of the system, we compare the new Vedalia system to the current Amazon system.
\begin{figure}[ht!]
\centering
    \includegraphics[height=0.4\textheight]{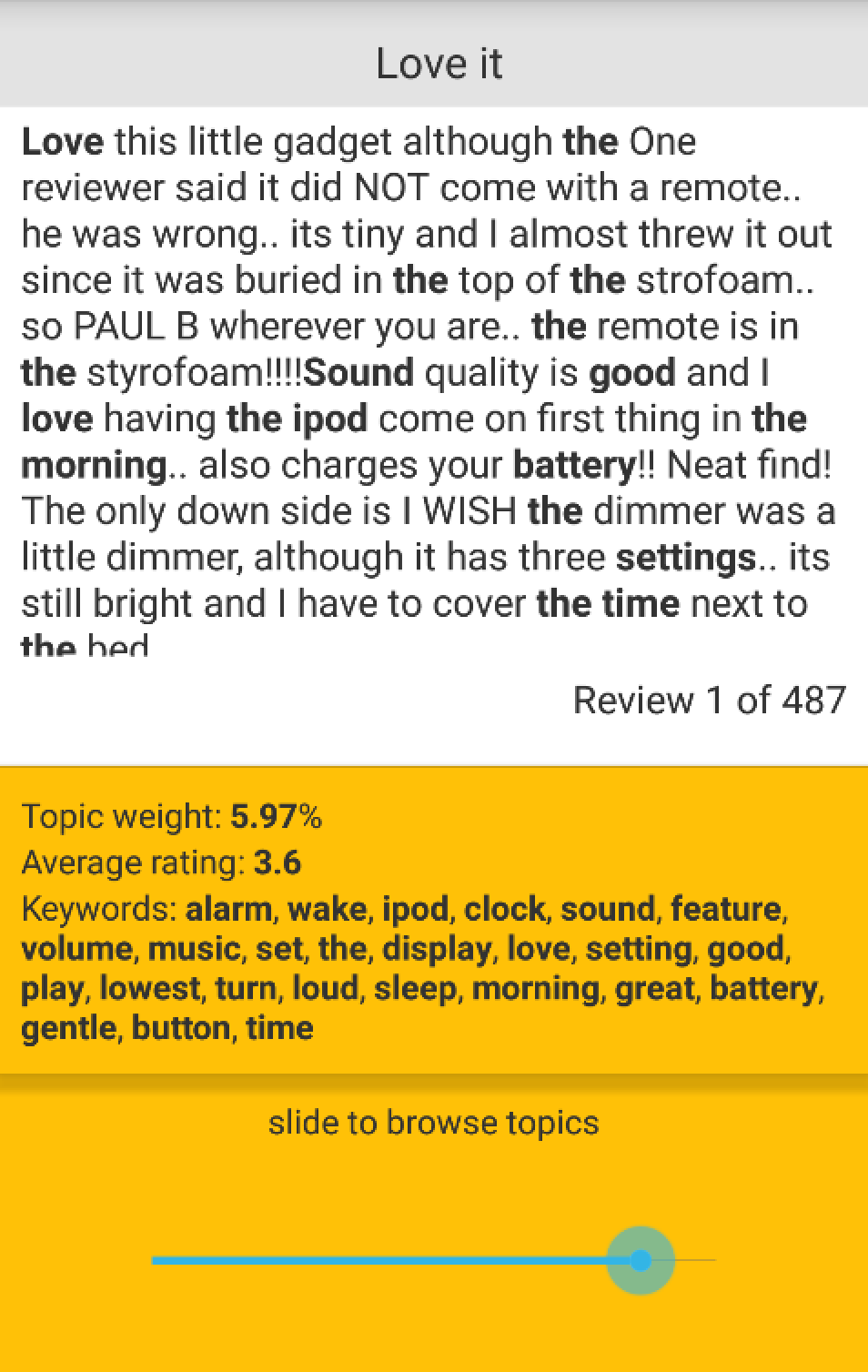}
    \caption{Above-average rating topic}
    \label{fig:good_topic}
\end{figure}
\begin{figure}[ht!]
\centering
    \includegraphics[height=0.4\textheight]{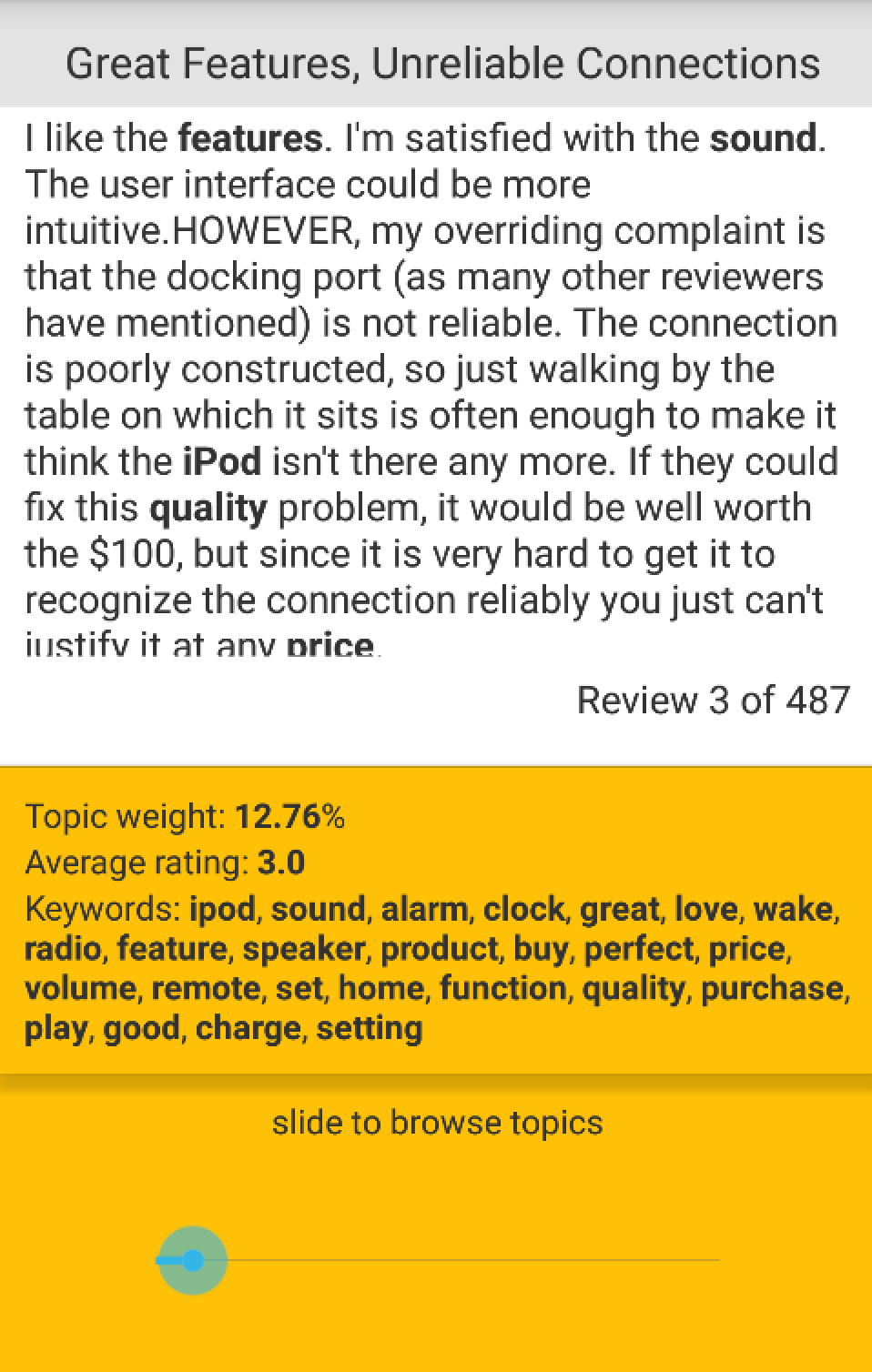}
    \caption{Below-average rating topic}
    \label{fig:bad_topic}
\end{figure}
\begin{figure}[ht!]
\centering
    \includegraphics[height=0.4\textheight]{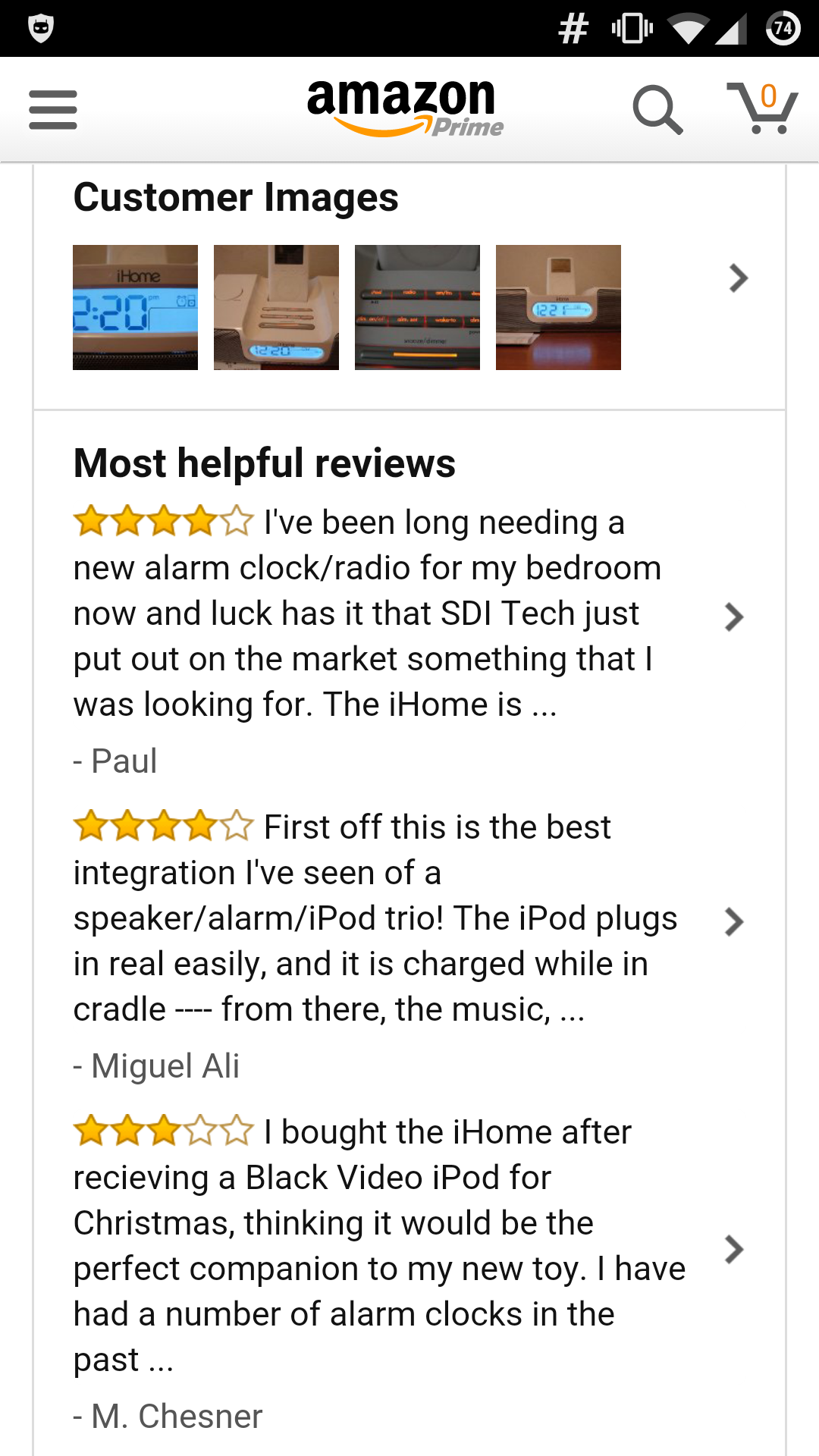}
    \caption{Review representation in Amazon app for Android}
    \label{fig:amazon_app}
\end{figure}

For our case study, we examine the use of the Quokka system for the iHome iH5 Clock Radio and Speaker System for iPod (ASIN B00080FO4O). At the time of modeling, the product had 487 reviews with an average rating of approximate 3.5 stars. Upon submitting a query for this product, the user is presented with the output given in Figures \ref{fig:good_topic} and \ref{fig:bad_topic}. In Figure \ref{fig:good_topic}, we see that the topic keywords are generally positive, with the highlighting bringing attention to the iHome's ability to charge your phone's battery and the mild (but not substantial) disappointment in the brightness of the screen when trying to sleep. In Figure \ref{fig:bad_topic}, we see more negative highlighted keywords, with emphasis on the product's sub-par build quality and unjustifiably high price.

In Figure \ref{fig:amazon_app}, we see the review representation in the official Amazon app for Android. Aside from the increased effort required to simply navigate through individual reviews, the system has no way of drawing the user's attention to any specific region of reviews, leaving the user to dig through mounds of text in order to find the specific information he is looking for.

For the iHome product modeled above, the time until initial results appeared was approximately 5 seconds, with final results appearing in 15 seconds.

\section{Future Work}
In the immediate future, we will be submitting a patch to MALLET \cite{MALLET} in order to fix the broken ParallelTopicModel parallel implementation. Many users of MALLET will note that the library runs substantially slower when using the multithreaded implementation. This slowdown is due to the ``Thread.sleep()'' calls in the inner loop of parameter estimation when using more than one thread, which appears to be a temporary hack used to correct for a concurrency bug (non-volatile thread-cached boolean value accessed from another thread). In designing our system, we corrected this bug and refactored the ParallelTopicModel and WorkerRunnable code in order to achieve substantially higher performance and fully utilize all cores in a multithreaded environment.

Regarding future work on RLDA, we wish to continually improve the model's performance on products with a limited number of reviews. The availability of a hierarchical structure of products allows for more advanced models that utilize product categories and reviews of similar products in order to better estimate topics in low-review situations. We also will be pursuing the idea of computing a single model per group of related products in order to leverage similarities in topics and improve topic estimation.

We would like to further investigate the performance of RLDA under some classical metrics to validate its superior performance compared to standard LDA in the context of product review modeling. We also plan to implement and test Chital at a larger scale and refine our user interface over the next few months so as to begin field testing with real users. As previously discussed, we will be releasing the app to the Google Play Store shortly -- as such, modeling performance, usability, and robustness in low-review situations are critical areas that require further optimization.
\bibliography{midway}
\bibliographystyle{icml2015}

\end{document}